\documentclass[10pt,twocolumn,letterpaper]{article}

\usepackage{cvpr}              

\usepackage{times}
\usepackage{epsfig}
\usepackage{graphicx}
\usepackage{subcaption}
\usepackage{amsmath}
\usepackage{amssymb}
\usepackage{multirow}
\usepackage{booktabs}
\usepackage[dvipsnames]{xcolor}
\usepackage{colortbl}
\newcommand{\stdvu}[1]{\scriptsize{\color{darkgray}(#1)}}

\usepackage{bigstrut}

\definecolor{tabhighlight}{HTML}{e5e5e5}

\usepackage{pifont}

\usepackage{multirow}
\usepackage{makecell}

\usepackage[dvipsnames]{xcolor}

\definecolor{cvprblue}{rgb}{0.21,0.49,0.74}
\usepackage[pagebackref,breaklinks,colorlinks,citecolor=cvprblue]{hyperref}

\title{BenchLMM: Benchmarking Cross-style Visual Capability of \\ Large Multimodal Models}

\author{Rizhao Cai$^{1}$\thanks{Equal contribution} , Zirui Song$^{2,3}$\textsuperscript{*},  Dayan Guan$^{1}$\thanks{Corresponding author} , Zhenhao Chen$^{4}$, Xing Luo$^{5}$, Chenyu Yi$^{1}$, Alex Kot$^{1}$ \\ 
$^{1}$Nanyang Technological University  $^{2}$University of Technology Sydney $^{3}$Northeastern University \\
$^{4}$Mohamed bin Zayed University of Artificial Intelligence $^{5}$Zhejiang University
}

\begin{document}
\maketitle

\begin{abstract}
Large Multimodal Models (LMMs) such as GPT-4V and LLaVA have shown remarkable capabilities in visual reasoning with common image styles. However, their robustness against diverse style shifts, crucial for practical applications, remains largely unexplored. In this paper, we propose a new benchmark, BenchLMM, to assess the robustness of LMMs against three different styles: artistic image style, imaging sensor style, and application style, where each style has five sub-styles. Utilizing BenchLMM, we comprehensively evaluate state-of-the-art LMMs and reveal: 1) LMMs generally suffer performance degradation when working with other styles; 2) An LMM performs better than another model in common style does not guarantee its superior performance in other styles; 3) LMMs' reasoning capability can be enhanced by prompting LMMs to predict the style first, based on which we propose a versatile and training-free method for improving LMMs; 4) An intelligent LMM is expected to interpret the causes of its errors when facing stylistic variations. We hope that our benchmark and analysis can shed new light on developing more intelligent and versatile LMMs. The benchmark and evaluation code have been released in \url{https://github.com/AIFEG/BenchLMM}.

\end{abstract}

\section{Introduction}
\label{sec:intro}
The dynamic landscape of computer vision has recently been shaped by the rise of Large Multimodal Models (LMMs). These models~\cite{liu2023visual,zhu2023minigpt4_v1,chen2023minigptv2,li2023otter,dai2023instructblip}, incorporating visual and textual data, have emerged as cornerstones in the quest for building general-purpose assistants for visual reasoning, a widely-studied task that involves reasoning and answering questions based on images. Their growing prominence and the enthusiasm surrounding visual reasoning are evident from numerous studies like GPT-4V~\citep{yang2023dawn} and LLaVA-1.5~\citep{liu2023improved}. As the interest in LMMs swells within the research community, understanding their depth, breadth, and limitations becomes imperative.

\begin{figure}[!t]
    \centering
        \includegraphics[page=1, width=\linewidth]{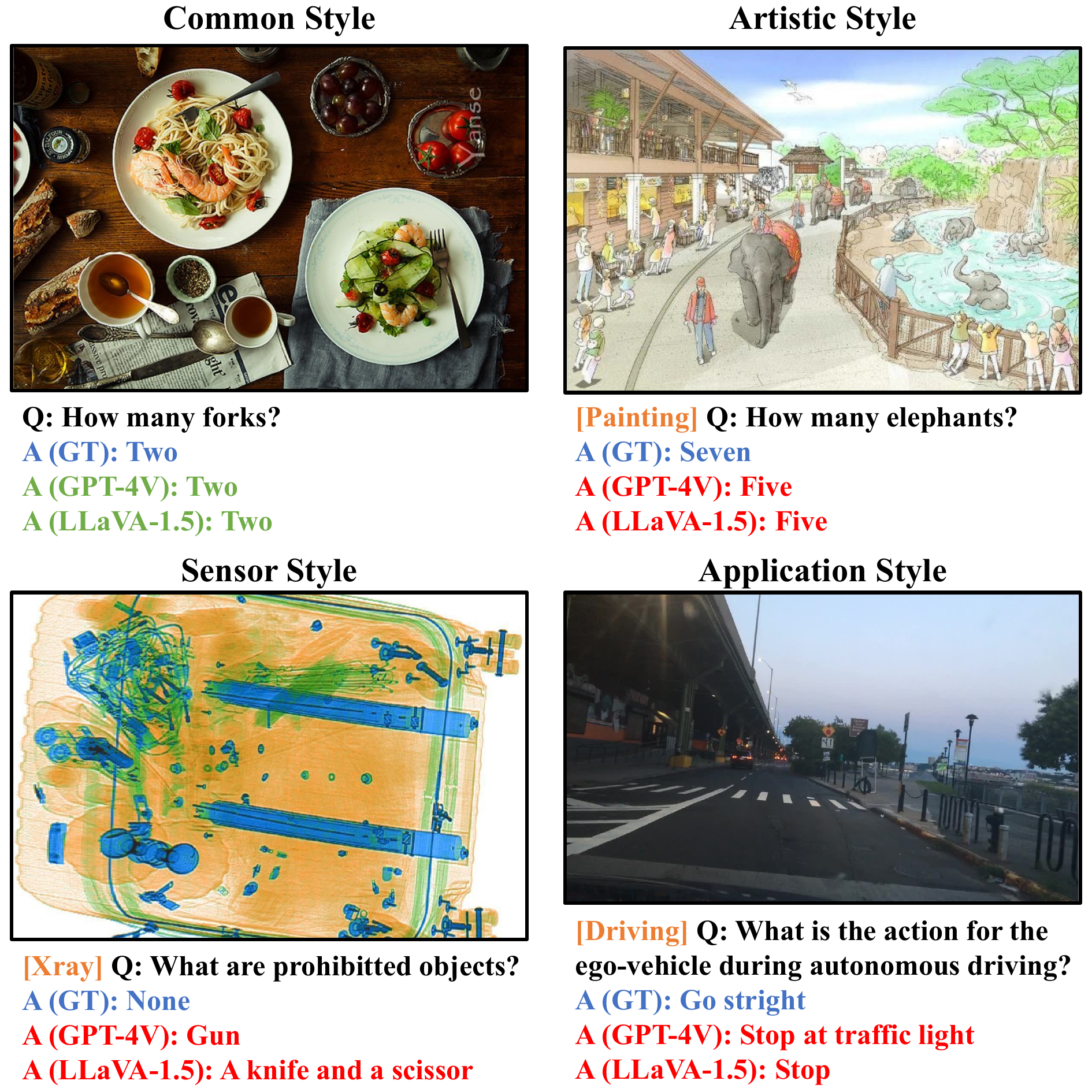}
        \vspace{-20pt}
        \caption{The motivation of our work. State-of-the-art LMMs, such as GPT-4V and LLaVA-1.5 can understand the image in a common style. However, these models have difficulties in reasoning images in other styles. }
        \vspace{-10pt}
    \label{fig_motivation}
\end{figure}

In pursuit of a comprehensive understanding of the visual reasoning capabilities of LLMs, numerous benchmarks~\cite{goyal2017making,liu2023mmbench,yu2023mmvet,li2023seed,li2023pope,fu2023mme} have been proposed to systematically evaluate the linguistic capabilities of LMMs. 
These linguistic capabilities are assessed through various types of responses for visual reasoning tasks, including Yes/No answers~\cite{fu2023mme}, multi-choice answers~\cite{li2023seed,liu2023mmbench}, single word or phrase answers~\cite{goyal2017making,hudson2019gqa,marino2019ok}.
However, while these benchmarks showcase textual diversity, they may fall short in terms of visual diversity as most images in these benchmarks are photographic images sourced from the internet. Despite providing valuable insights into LLMs' performance, there remains a gap in understanding how LLMs handle shifts in visual distribution. 
As illustrated in Figure~\ref{fig_motivation}, our preliminary study shows that LMMs tend to make mistakes in reasoning images with non-common styles.

To examine LMM's cross-style visual capability, we propose BenchLMM, a novel benchmark that systematically assesses existing LMMs in three distinct styles of distribution shifts: artistic style, sensor style, and application style. Firstly, there are images of different artistic styles. Objects in the common style are realistic, but objects can have other artistic styles, such as Cartoon, Painting, Sketch, etc.
Besides, we also identify the difference in sensor styles, as there are imaging sensors other than RGB cameras, such as infrared and x-ray, leading to the distribution shifts of the images. Moreover, we also identify the shift of knowledge required for different application styles. For example, the image recognition task requires knowledge about object shape and appearance, while the task of decision-making for robot action requires task-specific knowledge.

With BenchLLM, we evaluate cutting-edge LLMs through benchmark experiments and our designed error-reflection study. Through these comprehensive experiments,  we observe that these state-of-the-art LMMs perform well on the data of common style but suffer significant performance drops in the other styles; Also, we observe that an LMM achieving better performance in common style than another model does guarantee its superior performance in other styles; In addition, we find that prompting an LMM to predict what is the style of the data and then answer the question can enhance LMM's reasoning capability. Based on this, we propose Style Prompt Enhancement (SPE) as a training-free strategy to improve LMM's performance. 

Moreover, we analyze the failure case by conducting an error-reflection analysis by asking an LMM to interpret why it made mistakes in the Q\&A. We find that a more intelligent LMM can understand the detailed reason why it makes mistakes and can learn to derive the correct answer, which is an important capability that deserves attention when developing LMMs.
We hope that our benchmark and analysis can shed new light on developing more intelligent and versatile LMMs.  
Our contributions are summarized below:
\begin{itemize}
    \item We propose BenchLMM, the first benchmark that can be used to assess LMMs' capabilities, containing a diverse set of distribution shifts in artistical style, sensor style, and application style; 
    \item We broadly benchmark the existing large multimodal models, including commercial GPT-4V with quantitative metrics, which is the first time;
    \item Our in-depth benchmark analysis, our error-reflection study, and the proposed SPE provide a new aspect of understanding LMMs, which are useful for future research.
\end{itemize}

\section{Related works}

\noindent\textbf{Visual Reasoning.}
Visual reasoning, also known as visual question answering, aims to reason and answer questions about visual information~\cite{johnson2017clevr,antol2015vqa,johnson2017inferring}. Early developments ~\cite{fukui2016multimodal,kim2016hadamard,ben2017mutan} focus on designing more complex fusion mechanisms instead of simple summation fusion~\cite{antol2015vqa}. By including modern attention mechanism~\cite{vaswani2017attention}, numerous works~\cite{kim2018bilinear, nam2017dual,cadene2019murel,tan2019lxmert} contributed to the creation of transformer-based vision-language models. Recent methods have proposed to improve other aspects of visual question answering, including avoiding shortcut learning~\cite{dancette2021beyond,han2021greedy}, out-of-distribution generalization~\cite{cao2021linguistically,teney2021unshuffling}, 
addressing the linguistic bias~\cite{cao2021linguistically,teney2021unshuffling},
improving transformer-based vision-language models~\cite{yang2021auto,zhou2021trar}, external knowledge integration~\cite{ding2022mukea,gao2022transform}, and consistency regularization~\cite{shah2019cycle,tascon2023logical}. Lastly, the advent of large multimodal models has markedly enhanced the capabilities of visual reasoning. 

\vspace{6pt}
\noindent\textbf{Large Multimodal Models.}
Recent advancements in large language models (LLMs)~\cite{brown2020language,chowdhery2022palm,openai2023gpt,touvron2023llama} have inspired the development of large multimodal models (LMMs) that enhance LLMs with vision-language capabilities. 
Existing LMMs can be categorized into three main classes: multimodal instruction tuning~\cite{liu2023visual,liu2023improved,zhu2023minigpt4_v1,chen2023minigptv2,li2023otter,dai2023instructblip} that involves the finetuning of pre-trained LLMs
on multi-modal instruction-following data;  multimodal in-context learning~\cite{tsimpoukelli2021multimodal,li2023otter,yang2023mm,alayrac2022flamingo} that derives insights from few-shot task-specific samples; multimodal
chain-of-thought~\cite{zhang2023multimodal,ge2023chain} that encourages LMMs to articulate both the final answer and the underlying reasoning process.
Additionally, some commercial LMMs like Google's Bard~\cite{manyika2023overview} and Microsoft's GPT-4V~\cite{yang2023dawn}, with detailed information about their architecture remaining undisclosed, have shown impressive visual capabilities.

\vspace{6pt}
\noindent\textbf{Evaluation of LMMs.}
To evaluate their visual capabilities, existing LMMs have been tested on numbers of vision-language datasets. These datasets focus on specific visual reasoning tasks by designing different kinds of questions and answers, and the specific task includes visual recognition~\cite{goyal2017making,marino2019ok}, image description~\cite{chen2015microsoft,agrawal2019nocaps}, scene text understanding~\cite{singh2019towards,sidorov2020textcaps,ye2023mplug}, object hallucination evaluation~\cite{li2023evaluating}, complex reasoning~\cite{liu2023visual,yu2023mmvet}, and commonsense reasoning~\cite{zellers2019recognition,yu2022pacs}.
To facilitate a comprehensive comparison of LMMs, several benchmarks~\cite{fu2023mme,liang2021multibench,xu2023lvlm} have been established by encompassing various visual reasoning tasks with common-style images and diverse Q\&A pairs.
However,  the above benchmarks do not consider different styles of visual distribution shift, which motivates us to propose our BenchLMM.

\begin{figure*}[!t]
    \centering
        \includegraphics[page=1, width=\linewidth]{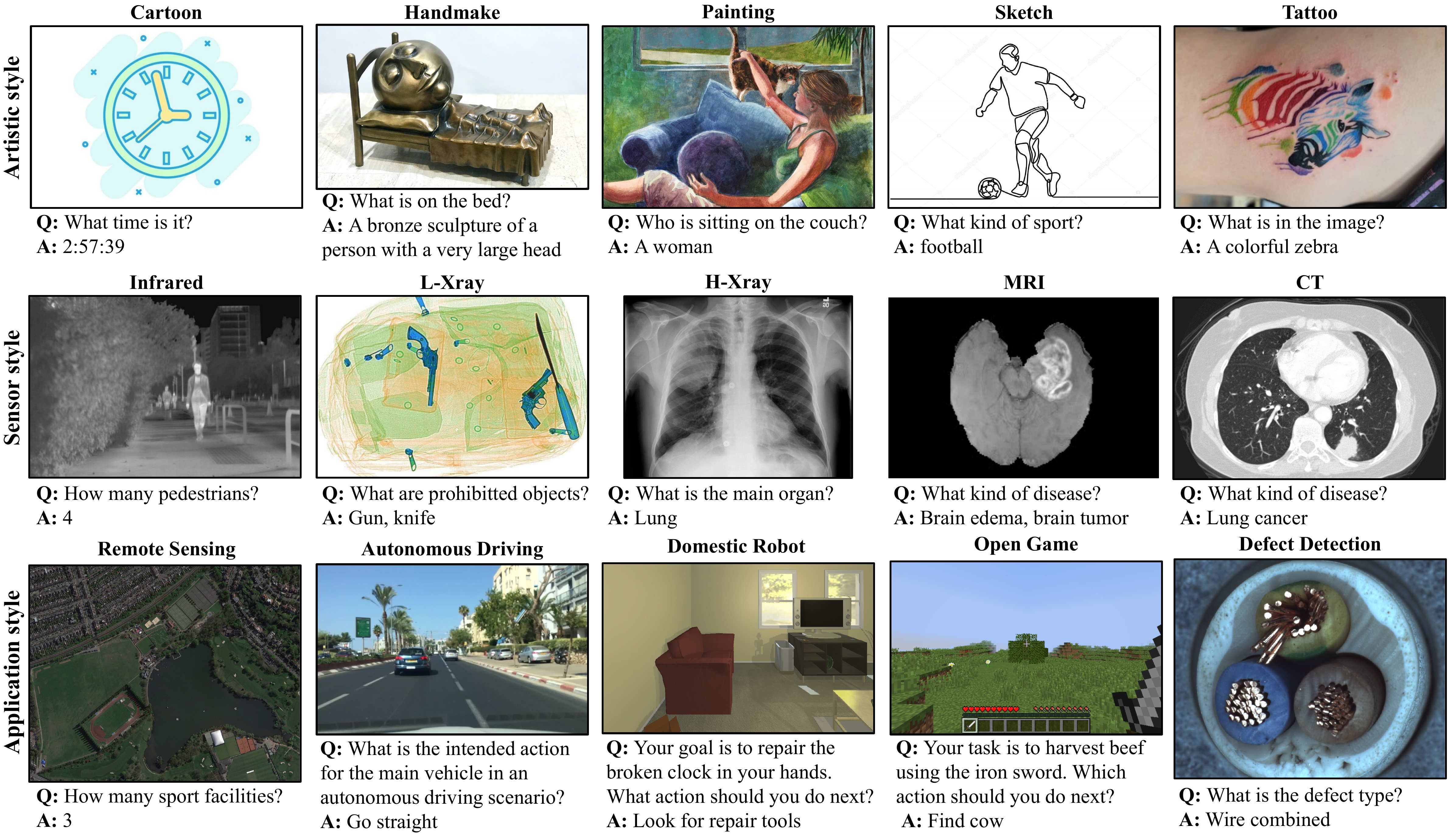}
        \vspace{-20pt}
        \caption{Benchmark examples. For each specific field, one example image and its corresponding Q\&A have been shown. Note that, for a simple presentation, the questions in \textit{Domestic Robot} and \textit{Open Game} have been simplified from multiple-choice format. Please see the Appendix for more examples and detailed questions.
        }
        \vspace{-10pt}
    \label{fig_benchmark_examples}
\end{figure*}

\section{BenchLMM} 
\subsection{Motivation}
We propose BenchLMM to investigate the cross-style capability of Large Multimodal Models (LMMs). This includes their proficiency in tasks like analyzing images with diverse styles, processing images acquired from non-RGB cameras, and interpreting images sourced from specific application knowledge. In this section, we will elaborate on how we build the benchmark that encompasses these diverse styles.

\textbf{Cross-artistic style}
Existing benchmarks for LMMs commonly use photo images for evaluation, which we name as `Common' style, because of their widespread use in applications such as education and e-commerce.
However, there are diverse image styles, including Cartoon, Sketch, etc. In Figure~\ref{fig_benchmark_examples}, we visually compare `Common' style images with various other styles. The object in other artistic styles have different distributions of object characteristics. For instance, a colorful zebra, common in painting and tattoo styles, is rare in `Common' style images.
While humans easily interpret such characteristics, the extent of LMMs' understanding of the other styles remains uncertain. This uncertainty arises from potential insufficient training on datasets encompassing these diverse styles. Therefore, we propose an evaluation to assess LMMs' robustness and proficiency in analyzing images across various artistic styles.

\textbf{Cross-sensor style.} 
Photographic images are typically obtained through RGB cameras for information sensing. However, alongside RGB cameras, other sensor types are also important and employed as complementary means of image capture. For instance, infrared cameras are frequently utilized in surveillance systems to capture images during nighttime, as RGB cameras rely on adequate illumination for optimal image quality. 
Additionally, X-ray sensors are used in airport security checks, enabling the detection of prohibited items within luggage.

These sensors feature distinct imaging processes compared to RGB cameras. Consequently, images acquired through various sensors exhibit unique intrinsic properties. The question of whether the performance of LMMs remains remarkable when applied to other sensors has not been thoroughly investigated. This issue will be addressed in our benchmark study.

\textbf{Cross-application style}
Existing LMMs are predominantly assessed in general application scenarios where questions can be answered using common knowledge~\cite{goyal2017making,fu2023mme}. For instance, identifying the object on a table in an image is a task that can be accomplished with general knowledge, and LMMs perform well in such cases. However, there exist specific application scenarios that demand specialized domain knowledge to address and solve. For example, when an autonomous driving control center processes the road image to decide the next actions: going straight, turning left, or turning right), specific driving knowledge is required. 

As existing LMMs are primarily evaluated with questions and answers (Q\&A) that rely on common knowledge, the extent to which an LMM can remain effective in situations requiring domain-specific knowledge has not been comprehensively explored. We also investigate this aspect in our proposed benchmark.

\subsection{Benchmark Construction}
We construct our benchmark by reorganizing publicly available datasets and relabeling them with VQA annotations.

\vspace{6pt}
\noindent\textbf{Artistic style data.} 
To construct a benchmark characterized by distributional shifts pertaining to artistic styles, we curated a diverse set of images encompassing Cartoon, Painting, Sketch, Handmade, and Tattoo styles from the COCO-O dataset \cite{mao2023coco}.  We systematically sampled 100 images for each stylistic category, resulting in a total of 500 images extracted from the COCO-O dataset. This dataset comprises a rich array of object types, providing a comprehensive evaluation ground for assessing the image understanding capabilities of the existing LMMs. Accordingly, the questions are prepared as asking the models to reason and answer questions about images, and the sample images with Q\&A are shown in the first row of Figure~\ref{fig_benchmark_examples}.

\begin{table*}[!t]
\centering
\caption{Evaluations of public LMMs on cross-style BenchLMM.
Note that Average$^\ast$ represents the average accuracy computed over five artistic-style benchmarks.
The best and the second best results are highlighted in \textbf{bold} and \underline{underline}, respectively.
All the numbers are presented in \% and the full score is 100\%.}
\label{tab:cross-style.}
\vspace{-10pt}
\small
\resizebox{\linewidth}{!}{
\begin{tabular}{@{}l|cc|c|ccccc|c@{}}
\toprule
            LMMs   &\makecell{Visual Encoder}  &\makecell{Language Model}   & Common    & Cartoon & Handmake & Painting & Sketch & Tattoo  & Average$^\ast$ \\ 
\midrule
GPT-4V~\cite{yang2023dawn} & - & - & \textbf{81.5} & \textbf{63.3} & \textbf{58.7} & \underline{58.2} & \textbf{64.7} & \textbf{68.3} & \textbf{62.6}  \\
\midrule
LLaVA-1.5-13B~\cite{liu2023improved} & CLIP-ViT-L/14 & Vicuna-13B & \underline{74.6} & \underline{62.0} & 56.6 & 57.6 & \underline{63.3} & 57.0 & \underline{59.3}  \\
InstructBLIP-13B~\cite{dai2023instructblip} & EVA-ViT-G & Vicuna-13B &72.7  & 59.0 & \underline{57.1} & \textbf{59.3} & 57.5 & \underline{61.4} & 58.9   \\
LLaVA-13B~\cite{liu2023visual} & CLIP-ViT-L/14 & Vicuna-13B &56.6 & 49.9 & 45.4 & 48.0 & 53.7 & 47.0 & 48.8   \\
MiniGPT4-13B~\cite{zhu2023minigpt4_v1} & CLIP-ViT-L/14 & Vicuna-13B & 56.9 & 57.8 & 28.4 & 37.8 & 42.6 & 32.3 & 40.0   \\
\midrule
LLaVA-1.5-7B~\cite{liu2023visual} & CLIP-ViT-L/14 & Vicuna-7B & 71.2 & 51.9 & 53.3 & 43.4 & 62.0 & 54.6 & 53.0   \\
InstructBLIP-7B~\cite{dai2023instructblip} & EVA-ViT-L & Vicuna-7B & 73.9 & 58.1 & 52.0 & 55.8 & 55.2 & 55.5 & 55.3  \\
MiniGPT4-v2-7B~\cite{chen2023minigptv2} & EVA-ViT-G & LLaMA2-7B & 66.4 & 28.7 & 37.8 & 40.9 & 41.7 & 28.6 & 35.5   \\
MiniGPT4-7B~\cite{zhu2023minigpt4_v1} & CLIP-ViT-L/14 & Vicuna-7B & 45.6 & 35.0 & 37.0 & 38.5 & 36.4 & 37.7 & 36.9   \\
Otter-7B~\cite{li2023otter} & CLIP-ViT-L/14 & MPT-7B & 64.8 & 45.1 & 49.0 & 39.8 & 44.1 & 42.4 & 44.1  \\
\bottomrule
\end{tabular}
}
\end{table*}

\renewcommand\arraystretch{0.9}
\begin{table*}[!t]
\centering
\caption{Evaluations of public LMMs on cross-sensor BenchLMM. Note that GPT-4V denied to test medical images (H-Xray/MRI/CT).
}\label{tab:cross-sensor}
\vspace{-10pt}
\small
\begin{tabular}{@{}l|cc|ccccc|c@{}}
\toprule
                LMMs   & \makecell{Visual Encoder} & \makecell{Language Model}    & Infrared & L-Xray & H-Xray & MRI & CT  & Average \\ 
\midrule
GPT-4V~\cite{yang2023dawn} & - & - & \textbf{55.0} & 50.0 & - & - & - & - \\
\midrule
LLaVA-1.5-13B~\cite{liu2023improved} & CLIP-ViT-L/14 & Vicuna-13B & 52.1 & \underline{51.5} & 55.9 & \underline{33.7} & 43.1 & \textbf{47.3} \\
InstructBLIP-13B~\cite{dai2023instructblip} & EVA-ViT-G & Vicuna-13B & 40.8 & 24.6 & \underline{56.6} & 20.9 & 43.4 & 37.3 \\
LLaVA-13B~\cite{liu2023visual} & CLIP-ViT-L/14 & Vicuna-13B 
&\underline{53.7}&46.0&39.1&28.4&39.7&41.4
\\
MiniGPT4-13B~\cite{zhu2023minigpt4_v1} & CLIP-ViT-L/14 & Vicuna-13B & 32.9 & 47.2 & 24.6 & 25.3 & 23.2 & 30.6 \\
\midrule
LLaVA-1.5-7B~\cite{liu2023visual} & CLIP-ViT-L/14 & Vicuna-7B & 46.5 & \textbf{53.2} & 49.4 & 29.6 & \textbf{46.8} & \underline{45.3} \\
InstructBLIP-7B~\cite{dai2023instructblip} & EVA-ViT-L & Vicuna-7B & 42.2 & 35.4 & \textbf{59.5} & 19.4 & 38.5 & 39.0 \\
MiniGPT4-v2-7B~\cite{chen2023minigptv2} & EVA-ViT-G & LLaMA2-7B & 30.1 & 43.0 & 31.5 & 17.4 & 17.8 & 28.2 \\
MiniGPT4-7B~\cite{zhu2023minigpt4_v1} & CLIP-ViT-L/14 & Vicuna-7B & 35.1 & 44.1 & 23.7 & 16.8 & 15.3 & 27.0 \\
Otter-7B~\cite{li2023otter} & CLIP-ViT-L/14 & MPT-7B & 30.2 & 46.4 & 39.1 & \textbf{47.8} & \underline{46.3} & 42.0 \\
\bottomrule
\end{tabular}
\end{table*}

\vspace{6pt}
\noindent\textbf{Sensor-style data.}
Other than RGB cameras, Infrared sensors, X-ray sensors, Magnetic Resonance Imaging (MRI) sensors, and Computed Tomography (CT) sensors are often used to collect image data for human beings to perceive and analyze.  Our benchmark examines whether the LMMs can still have the comparative reasoning capability toward these images that are intrinsically different from RGB photo images.
In the case of Infrared data, we collected a set of 200 images from the datasets that focus on street scenes analysis toward with pedestrians, buildings, and vehicles \cite{2015infra-pedestrian,2017-infrared-vehicle,2021infra-lowlight,2022-infrared-detect}. Questions for LMMs mainly revolve around analyzing the contents of these street scenes. Specifically, we design some thermal-related questions such as determining the hottest content.  
An illustrative example is provided in the second row of Figure~\ref{fig_benchmark_examples}.

For Low-energy X-ray (L-Xray)
data from the SiXray-D dataset \cite{nguyen2022towards}, we gathered 200 images, including 160 with prohibited objects (knife, scissors, wrench, gun, hammer) and 40 without any prohibited objects. Questions are designed to assess the ability to detect security risks in X-ray images, such as identifying forbidden objects. A sample question is presented in the second row of Figure~\ref{fig_benchmark_examples}.
In the medical application, high-energy X-ray (H-Xray)
, MRI, and CT scanning images (200 each) were extracted from the \cite{isbi}. Questions for LMMs in this context involve tasks like recognizing organs or identifying diseases in the respective images. 
By evaluating and comparing inference results between different sensor types, our benchmark aims to uncover challenges associated with cross-sensor analysis.

\vspace{6pt}
\noindent\textbf{Application-style data.}
In order to establish a comprehensive cross-application-style benchmark, we systematically integrated diverse task domains into our evaluation framework. We consider four representative applications: remote sensing, autonomous driving, agent action prediction, and defect detection.
Specifically, for the remote sensing application, we utilized 200 images sourced from the DOTA dataset \cite{xia2018dota} and annotated them in the VQA format from a satellite perspective.
For the autonomous driving task, we extracted 250 images from the BDD100K dataset \cite{BDD100K}, featuring various weather conditions. Our primary focus was on designing questions and answers aimed at predicting driving actions such as going straight, turning left, or turning right. 
An illustrative example is provided in the third row of Figure~\ref{fig_benchmark_examples}.

To introduce the agent action prediction application, we meticulously selected 100 images from the Domestic Robot dataset \cite{chen2023towards} and an additional 117 images from the Open Game dataset \cite{chen2023towards}. The associated questions center around predicting the next action to be taken under the current circumstances. Furthermore, for evaluating defect detection capabilities, we curated 100 images from the Defect Detection dataset \cite{MVTec}, with questions focused on identifying defect types and determining the presence of defects. This diverse set of applications and datasets collectively forms a diverse benchmark, allowing for a comprehensive assessment of the proposed approach across various application domains.
 
\subsection{
Style Prompt Enhancement
}
It is generally expected that providing additional prior information to LMMs' inputs can enhance performance. Nevertheless, such contextual priors, such as knowledge of artistic styles, might be unavailable in automated inference scenarios.
We are inspired by human cognitive behavior that humans often comprehend the style of an image before delving into object recognition, akin to appreciating a painting by Van Gogh, where recognition of the impressionist style precedes identification of objects within the artwork.
Other contextual priors, like sensor or application styles, can also enhance LMMs' performance. 
For example, identifying people in infrared images is easier when LMMs know that the images are captured using infrared thermal cameras, since humans typically show consistent temperature.

With this insight, we investigate the efficacy of prompting LMMs to predict the style of an image before addressing associated questions. To this end, we devised a Style Prompt Enhancement (SPE) by introducing a text prompt, instructing the model to ``Determine the style of the image before answering the question." 
SPE aims to leverage stylistic cues as a precursor to visual question answering, potentially enhancing the model's ability of visual reasoning.
Despite its simplicity, SPE's versatility allows it to be applied directly to images of various styles without any additional prompt tuning or model training, clearly boosting the overall performance of LMMs, as detailed in Section~\ref{Experiments_PE}.

\section{Experiment}\label{sec-exp}

\renewcommand\arraystretch{0.9}
\begin{table*}[!t]
\centering
\caption{
Evaluations of public LMMs on cross-task BenchLMM.
}\label{tab:cross-task}
\vspace{-10pt}
\small
\begin{tabular}{@{}l|cc|ccccc|c@{}}
\toprule
LMMs   & \makecell{Visual Encoder} & \makecell{Language Model} & \makecell{Remote \\Sensing} & \makecell{Autonomous \\Driving} & \makecell{Domestic \\Robot} & \makecell{Open \\Game} & \makecell{Defect \\Detection}  & \makecell{Average} \\ 
\midrule
GPT-4V~\cite{yang2023dawn} & - & - & \textbf{69.7} & \underline{35.8} & \underline{56.0} & \textbf{74.0} &\textbf{64.4} &\textbf{60.0} \\
\midrule
LLaVA-1.5-13B~\cite{liu2023improved} & CLIP-ViT-L/14 & Vicuna-13B & \underline{65.6} & 28.6 & 49.0 & 21.4 & \underline{60.3} & \underline{45.2} \\
InstructBLIP-13B~\cite{dai2023instructblip} & EVA-ViT-G & Vicuna-13B 
&63.6&32.6&32.0&19.5&47.0&38.9
\\
LLaVA-13B~\cite{liu2023visual} & CLIP-ViT-L/14 & Vicuna-13B &38.7 &31.7 &\textbf{59.0} &\underline{43.8} &28.2 &40.3 \\
MiniGPT4-13B~\cite{zhu2023minigpt4_v1} & CLIP-ViT-L/14 & Vicuna-13B & 33.3 & 22.6 & 48.0 & 34.2 & 31.7 & 34.2 \\
\midrule
LLaVA-1.5-7B~\cite{liu2023visual} & CLIP-ViT-L/14 & Vicuna-7B & 61.7 & 31.3 & 44.0 & 15.4 & 59.6 & 42.2 \\
InstructBLIP-7B~\cite{dai2023instructblip} & EVA-ViT-L & Vicuna-7B & 63.5 & \textbf{36.7} & 42.0 & 24.8 & 30.8 & 39.6 \\
MiniGPT4-v2-7B~\cite{chen2023minigptv2} & EVA-ViT-G & LLaMA2-7B & 34.5 & 17.6 & 40.0 & 22.2 & 18.8 & 26.6 \\
MiniGPT4-7B~\cite{zhu2023minigpt4_v1} & CLIP-ViT-L/14 & Vicuna-7B & 27.9 & 20.8 & 41.0 & 40.1 & 13.6 & 28.7 \\
Otter-7B~\cite{li2023otter} & CLIP-ViT-L/14 & MPT-7B & 44.8 & 29.7 & 36.0 & 18.8 & 27.3 & 31.3 \\
\bottomrule
\end{tabular}
\end{table*}

\begin{figure*}[!t]
    \centering
        \includegraphics[page=1, width=\linewidth]{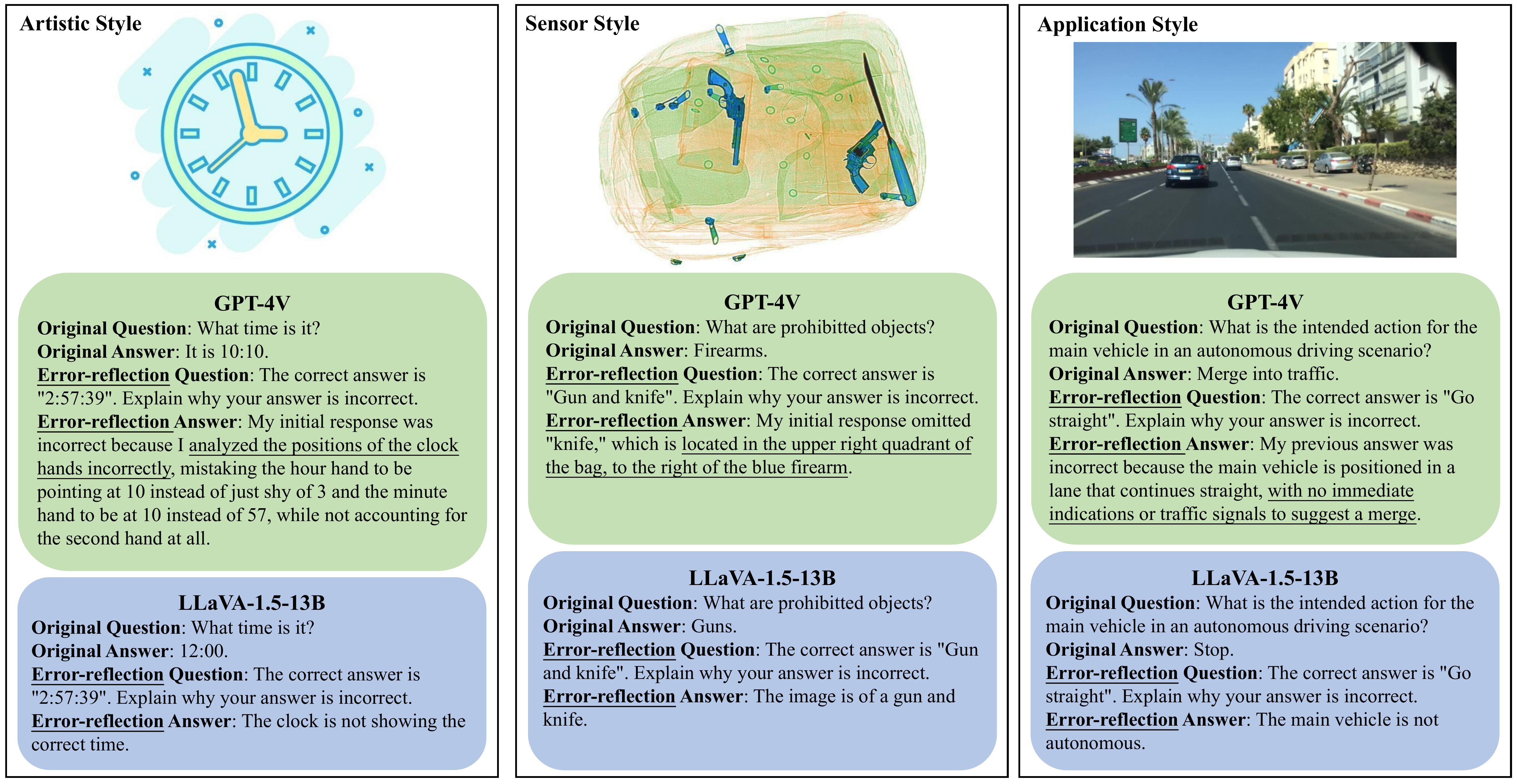}
        \vspace{-20pt}
        \caption{Error-reflection capability comparisons: Unlike LLaVA-1.5-13B, GPT-4V possesses the ability to analyze its errors when provided with the correct answers. }
    \label{fig_failure_case}
\end{figure*}

\renewcommand\arraystretch{0.9}
\begin{table*}[!t]
\centering
\caption{Results of our proposed SPE on the BenchLMM benchmark}\label{tab:SPE-bench}
\vspace{-10pt}
\small
\begin{tabular}{@{}l|ccccc|c@{}}
\toprule
            LMMs       & Cartoon & Handmake & Painting & Sketch & Tattoo  & Average \\ 
\midrule
LLaVA-1.5-13B~\cite{liu2023improved}       &62.0&56.6&57.6&63.3&57.0&59.3 \\
+ \textbf{SPE (Ours)} &\textbf{67.6}&\textbf{57.7}&\textbf{67.9}&\textbf{68.8}&\textbf{63.5}&\textbf{65.1}~\stdvu{{+5.8}}  \\
\midrule
InstructBLIP-13B~\cite{dai2023instructblip}   &59.0&57.1&59.3&57.5&61.4&58.9  \\
+ \textbf{SPE (Ours)} &\textbf{61.1}&\textbf{62.3}&\textbf{64.8}&\textbf{60.9}&\textbf{63.3}&\textbf{62.5}~\stdvu{{+3.6}}  \\
\midrule
\midrule
                LMMs       & Infrared & L-Xray & H-Xray & MRI & CT  & Average \\ 
\midrule
LLaVA-1.5-13B~\cite{liu2023improved}       &52.1&51.5&55.9&\textbf{33.7}&43.1&47.3 \\
+ \textbf{SPE (Ours)} &\textbf{58.7}&\textbf{62.8}&\textbf{56.2}&33.6&\textbf{47.7}&\textbf{51.8}~\stdvu{{+4.5}} \\
\midrule
InstructBLIP-13B~\cite{dai2023instructblip}   &40.8&24.6&56.6&20.9&\textbf{43.4}&37.3  \\
+ \textbf{SPE (Ours)} &\textbf{43.5}&\textbf{24.8}&\textbf{57.7}&\textbf{25.0}&42.6&\textbf{38.7}~\stdvu{{+1.4}}  \\
\midrule
\midrule
LMMs   & \makecell{Remote \\Sensing} & \makecell{Autonomous \\Driving} & \makecell{Domestic \\Robot} & \makecell{Open \\Game} & \makecell{Defect \\Detection}  & \makecell{Average} \\
\midrule
LLaVA-1.5-13B~\cite{liu2023improved}       &65.6&28.6&49.0&21.4&60.3&45.2   \\
+ \textbf{SPE (Ours)} &\textbf{68.0}&\textbf{29.4}&\textbf{51.0}&\textbf{22.7}&\textbf{61.7}&\textbf{46.6}~\stdvu{{+1.4}} \\
\midrule
InstructBLIP-13B~\cite{dai2023instructblip}   &\textbf{63.6}&32.6&\textbf{32.0}&\textbf{19.5}&47.0&38.9  \\
+ \textbf{SPE (Ours)} &63.5&\textbf{38.8}&31.0&19.0&\textbf{49.4}&\textbf{40.3}~\stdvu{{+1.4}}  \\
\bottomrule
\end{tabular}
\end{table*}

\subsection{Experimental Setup}

In this paper, we gather and assess a selection of existing open-source LMMs, which include LLaVA \cite{liu2023visual, liu2023improved}, MiniGPT4 \cite{chen2023minigptv2,zhu2023minigpt4_v1}, InstructBLIP~\cite{dai2023instructblip}, Otter~\cite{li2023otter}, and OpenFlamingo-7B~\cite{awadalla2023openflamingo}. We initialize these models with pre-trained weights and conduct evaluations using our dataset. In addition, we use the variants of these models, where the language encoders and vision encoders are changed to investigate how different encoders have impacts on the cross-field capabilities. Moreover, we acknowledge the recent release of the latest commercial LMM, GPT-4V. We perform a quantitative evaluation of GPT-4V with our dataset and compare its performance with the aforementioned open-source LMMs, and such a quantitative evaluation is the first time.

In our performance evaluation, we adhere to previous works \cite{liu2023visual,liu2023improved} methodology, employing the ChatGPT API to gauge the proximity of answers predicted by the LMMs to ground-truth answers. Here, we designate the output of the LMM as $y$, and the ground-truth label as $\hat{y}$. The evaluation utilizes the prompt ``Compare the similarity between {$y$} and {$\hat{y}$} with a correctness score ranging from 0.0 (totally wrong) to 1.0 (totally right). The middle score provides the percentage of correctness.''

\subsection{Evaluations of public LMMs}

\subsubsection{Cross-artistic capability}
In most existing works~\cite{goyal2017making,liu2023mmbench,yu2023mmvet,li2023seed,li2023pope,fu2023mme}, LMMs are predominantly evaluated using images in the 'Photo' style, leading to a gap in understanding their performance across diverse artistic styles. We extend the evaluation scope by examining LMMs' performance with various artistic styles beyond the common 'Photo' style. Results, as detailed in Table~\ref{tab:cross-style.}, reveal a notable decline in LMMs' effectiveness when processing these artistic styles. This trend suggests a potential overfitting of LMMs to the 'Photo' style, highlighting their limited adaptability to varied artistic styles, a capability that humans typically possess. Interestingly, GPT-4V, despite being a robust commercial model, exhibits similar limitations in handling diverse styles.

Besides, LLaVA-1.5-13B \cite{liu2023improved} is improved from LLaVA-13B \cite{liu2023visual} by introducing more training tricks and data, and LLaVA-1.5-13B outperform LLaVA-13B by a clear margin on the `Common' and other artistic styles. This comparison reveals the importance of data and training strategy as LLaVA-1.5-13B and LLaVA-13B use the same visual encoder and language encoder. However, we also find that a better performance on `Common' does not necessarily indicate a better performance on other styles. 
For example, compared with InstructBLIP-7B, InstructBLIP-13B demonstrates better performance across five artistic-style benchmarks in terms of average accuracy, though it shows comparatively lower performance in the `Common' style.
Therefore, such results convey that it is important to conduct comprehensive evaluations to fully validate an LMM's capability rather than only focus on the `Common' style, which shows the importance of our proposed benchmark. 

\subsubsection{Cross-sensor capability} 
We benchmark the LMM models' cross-sensor capability in Table~\ref{tab:cross-sensor}. 
Because GPT-4V denied to provide medical diagnostics or identify potential abnormalities in medical images like H-Xray/MRI/CT, GPT-4V is only evaluated on the Infrared and L-Xray sensor data, and achieves the results are 55.0\% and 50.0\% respectively, which are far lower than the `Common' style (81.5\%). 
Meanwhile, other LMM models' performance on images captured by X-ray, MRI, CT, and Infrared dropped significantly compared to the performance on `Common' captured by the RGB camera sensor. Therefore, the intrinsic difference in the imaging process between RGB cameras and other sensors is also a crucial challenge to LMMs.    

With respect to the MRI, H-Xray, and CT images, the questions may require specific medical knowledge to answer, and it may be argued that the performance drop is due to the knowledge gap. It is also observed from the L-Xray, where the objects (\ie, knife, gun, hammer, etc.) can be recognized with only common knowledge, but the LMMs' performance on the L-Xray data is significantly lower than `Common', which further validates the challenge brought by the sensor difference.

\subsubsection{Cross-application capability}
In Table~\ref{tab:cross-task}, we evaluate the different LMMs on various applications, finding  GPT-4V generally outperforms others, except in autonomous driving where its accuracy drops to 35.8\%. This decrease is likely due to the challenges in the Autonomous Driving dataset, such as recognizing low-resolution traffic signs and requiring specific driving knowledge, where both GPT-4 and open-source LMMs have poor performance.
We also observe that LLaVA-1.5-13B achieves comparable performance as GPT-4V on the Remote Sensing and Defect Detection datasets, which seems that they have comparable cross-task capability. However, we surprisingly find that LLaVA-1.5-13B merely achieves 21.4\% but GPT-4V achieves up to 74.0\%. As GPT-4V is close-source, the technical details are not disclosed and we conjecture that the training data for GPT-4V may involve data similar to the Open Game dataset.

\subsubsection{Summary of the BenchLMM}
As we can summarize from the above experiments, 1) existing LMMs generally suffer performance degradation when processing cross-style images.
2) Achieving better performance on the `Common' data,  which is of majority, does not guarantee better performance on other styles. 3) It is necessary to conduct a comprehensive evaluation of different dimensions because in some tasks, such as medical or security, the error information will lead to a significant loss.

\renewcommand\arraystretch{0.9}
\begin{table*}[!t]
\centering
\caption{Results of our proposed SPE on style-transfer images}
\label{tab:SPE-style}
\vspace{-10pt}
\small
\begin{tabular}{@{}l|ccccc|c@{}}
\toprule
               LMMs    & Cartoon & Handmake & Painting & Sketch & Tattoo  & Average \\ 
\midrule
LLaVA-1.5-13B~\cite{liu2023improved}   &62.1&61.1&52.4&68.8&41.2&57.1 \\
+ \textbf{SPE (Ours)} &\textbf{67.4}&\textbf{62.1}&\textbf{59.4}&\textbf{69.4}&\textbf{45.5}&\textbf{60.8}~\stdvu{{+3.7}}\\
\midrule
InstructBLIP-13B~\cite{dai2023instructblip}    & 69.3 & 70.8 & 61.3 & 53.8 & 65.0 & 64.0 \\
+ \textbf{SPE (Ours)} & \textbf{73.0} & \textbf{71.6} & \textbf{64.2} & \textbf{62.1} & \textbf{65.3} & \textbf{67.2}~\stdvu{{+3.2}} \\
\bottomrule
\end{tabular}
\end{table*}

\begin{figure*}[!h]
    \centering
        \includegraphics[page=1, width=\linewidth]{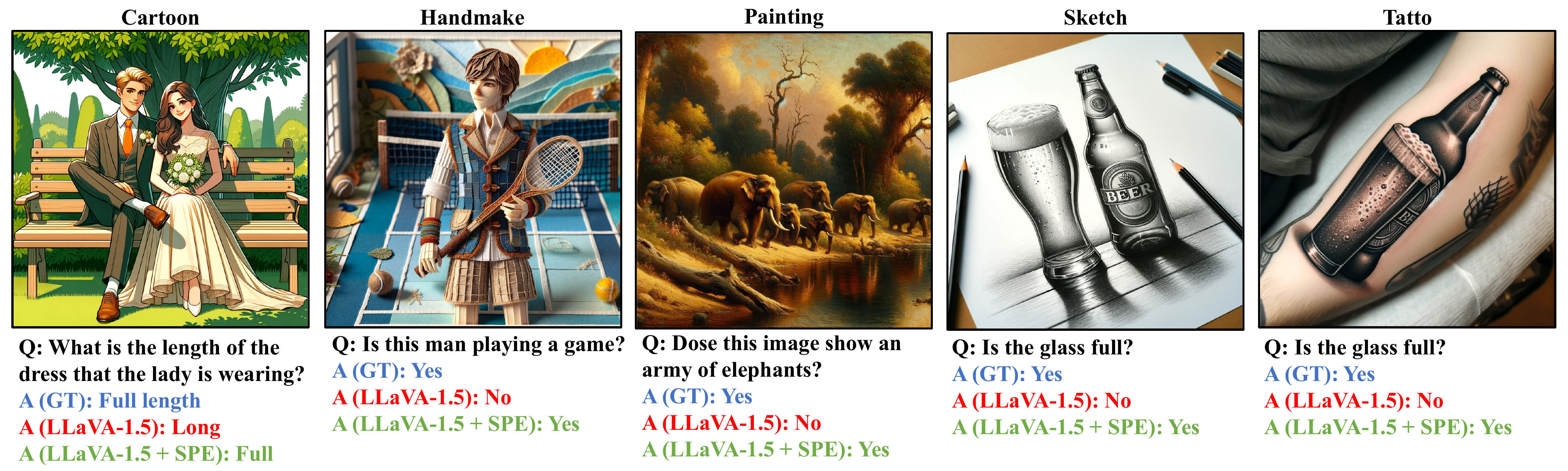}
        \vspace{-20pt}
        \caption{Qualitative comparisons between LLaVA-1.5-13B and LLaVA-1.5-13B + SPE (Ours) on style-transfer images.
        }
    \label{fig_SPE}
\end{figure*}

\subsubsection{Error-reflection Capability}
In order to understand the LMMs further, we let the LMM do an error-reflection that uses itself to parse the error information. When a model gives wrong answers, we warp up the conversation: the previous question input and predicted answer output. 
Specifically, we add ``The correct answer is [GT Anwser]. Explain why your answer is incorrect. Use a single sentence." to the conversation, which will be the new input of LMMs. 

Although both GPT-4V and LLaVA-1.5-13B produce incorrect answers, GPT-4V demonstrates the ability to reason the correct answer through a detailed inference process. 
As illustrated in Figure~\ref{fig_failure_case}, GPT-4V can offer insightful analysis beyond the ground-truth answer, including the positions of clock hands for estimating time, the location of prohibited objects, and the traffic signals for driving planning.

By contrast, the error-reflection capability of LLaVA-1.5-13B is inferior to that of GPT-4V, as it merely rephrases the information from the correct answers provided. 
Therefore, we reveal that another important capability of a large multimodal model is whether the model can reflect how to derive the correct answer. 
Studying such capability is useful to common users. For example, when users use LMMs for self-learning, the user can provide reference answers and LMMs can provide a more detailed inference process, which can improve the learning effect. Moreover, the answer of GPT-4V can also be useful materials used to train open-source LMMs in the future.

\subsection{Experiments of Style Prompt Enhancement}
\label{Experiments_PE}
In this section, we aim to study if a large multimodal model can be enhanced by prompting the model to estimate the style first and then answer the question, based on which we propose Style Prompt Enhancement (SPE) method and conduct experiments with with our BenchLMM. 
Specifically, we examined our SPE with LLaVA-1.5-13B~\cite{liu2023improved} and InstructBLIP-13B~\cite{dai2023instructblip} because these two models are representative with better average performance than MiniGPT4~\cite{zhu2023minigpt4_v1} and Otter~\cite{li2023otter}.

We conduct experiments on our BenchLLM with the data in different artistic, sensor, and application styles, and the results are presented in Table~\ref{tab:SPE-bench}. 
Our proposed SPE can benefit the LMMs by achieving general improvement over different styles, which means that reasoning the style first is an effective prompting method to improve LMMs' visual reasoning capability.
Besides, in the experiments of crossing to different applications (Remote Sensing, Autonomous Driving, etc.), we observe that its benefit is smaller than others. For example, LLaVA-1.5-13B gains 5.8\% and 4.5\% improvement in the experiments of cross-artistic and cross-sensor styles, while the improvement is merely 1.4\% when crossing applications. We conjecture that specific application knowledge could inferred from only the style, which limits the benefits of SPE. 
Moreover, we observe that LLaVA-1.5-13 generally obtains more benefits than InstructBLIP-13B, which means that the stronger the LMM, and more benefits from SPE, which is because a stronger LMM has stronger inference capability and can better utilize prior information.

We employ the latest commercial style transfer model, DALL-E3~\cite{shi2020improving}, to transfer the `Common' style images used in Table~\ref{tab:cross-style.} to different artistic styles (\ie, cartoon, handmake, painting, sketch, and tattoo). 
In Figure~\ref{fig_SPE}, we provide visual examples of images in these different styles. These images share similar structures and layouts between different styles. As shown in the Table~\ref{tab:SPE-style}, the performance of the transferred artistic styles also shows significant degradation, which further validates that LMMs suffer from difficulties in recognizing images of different styles. Furthermore, our SPE significantly enhances the performance of LMMs in this challenging task.

\section{Conclusion}
In this work, we introduce BenchLMM, a novel benchmark for quantitatively evaluating large multimodal models (LMMs) across varied visual distribution shifts, including artistic, sensor, and application styles. Our evaluations on numerous existing LMMs highlight two key findings: 1) LMMs typically under-perform with non-common image styles. 2) Our style prompt enhancement approach, inspired by human perception of art, significantly improves LMMs' visual reasoning without extra fine-tuning. Additionally, our error-reflection study shows that stronger LMMs can self-diagnose errors by providing insights not given by humans, whereas less capable models merely restate correct answers without new insights. Our benchmark offers a comprehensive tool for LMM research, emphasizing the importance of error-reflection capabilities in future developments.

\paragraph{Acknowledgement.} 
This research is supported in part by the Rapid-Rich Object Search (ROSE) Lab of Nanyang Technological University and the NTU-PKU Joint Research Institute (a collaboration between NTU and Peking University that is sponsored by a donation from the Ng Teng Fong Charitable Foundation). 
We are deeply grateful to Yaohang Li from the University of Technology Sydney for his invaluable assistance in conducting the experiments, and to Jingpu Yang, Helin Wang, Zihui Cui, Yushan Jiang, Fengxian Ji, and Yuxiao Hang from NLULab@NEUQ (Northeastern University at Qinhuangdao, China) for their meticulous efforts in annotating the dataset. We also would like to thank Prof. Miao Fang (PI of NLULab@NEUQ) for insightful suggestion during discussion on this project.

{
    \small
    \bibliographystyle{ieeenat_fullname}
    \bibliography{main}

\begin{thebibliography}{69}
\providecommand{\natexlab}[1]{#1}
\providecommand{\url}[1]{\texttt{#1}}
\expandafter\ifx\csname urlstyle\endcsname\relax
  \providecommand{\doi}[1]{doi: #1}\else
  \providecommand{\doi}{doi: \begingroup \urlstyle{rm}\Url}\fi

\bibitem[Agrawal et~al.(2019)Agrawal, Desai, Wang, Chen, Jain, Johnson, Batra, Parikh, Lee, and Anderson]{agrawal2019nocaps}
Harsh Agrawal, Karan Desai, Yufei Wang, Xinlei Chen, Rishabh Jain, Mark Johnson, Dhruv Batra, Devi Parikh, Stefan Lee, and Peter Anderson.
\newblock Nocaps: Novel object captioning at scale.
\newblock In \emph{Proceedings of the IEEE/CVF international conference on computer vision}, pages 8948--8957, 2019.

\bibitem[Alayrac et~al.(2022)Alayrac, Donahue, Luc, Miech, Barr, Hasson, Lenc, Mensch, Millican, Reynolds, et~al.]{alayrac2022flamingo}
Jean-Baptiste Alayrac, Jeff Donahue, Pauline Luc, Antoine Miech, Iain Barr, Yana Hasson, Karel Lenc, Arthur Mensch, Katherine Millican, Malcolm Reynolds, et~al.
\newblock Flamingo: a visual language model for few-shot learning.
\newblock \emph{Advances in Neural Information Processing Systems}, 35:\penalty0 23716--23736, 2022.

\bibitem[Antol et~al.(2015)Antol, Agrawal, Lu, Mitchell, Batra, Zitnick, and Parikh]{antol2015vqa}
Stanislaw Antol, Aishwarya Agrawal, Jiasen Lu, Margaret Mitchell, Dhruv Batra, C~Lawrence Zitnick, and Devi Parikh.
\newblock Vqa: Visual question answering.
\newblock In \emph{Proceedings of the IEEE international conference on computer vision}, pages 2425--2433, 2015.

\bibitem[Awadalla et~al.(2023)Awadalla, Gao, Gardner, Hessel, Hanafy, Zhu, Marathe, Bitton, Gadre, Sagawa, Jitsev, Kornblith, Koh, Ilharco, Wortsman, and Schmidt]{awadalla2023openflamingo}
Anas Awadalla, Irena Gao, Josh Gardner, Jack Hessel, Yusuf Hanafy, Wanrong Zhu, Kalyani Marathe, Yonatan Bitton, Samir Gadre, Shiori Sagawa, Jenia Jitsev, Simon Kornblith, Pang~Wei Koh, Gabriel Ilharco, Mitchell Wortsman, and Ludwig Schmidt.
\newblock Openflamingo: An open-source framework for training large autoregressive vision-language models.
\newblock \emph{arXiv preprint arXiv:2308.01390}, 2023.

\bibitem[Ben-Younes et~al.(2017)Ben-Younes, Cadene, Cord, and Thome]{ben2017mutan}
Hedi Ben-Younes, R{\'e}mi Cadene, Matthieu Cord, and Nicolas Thome.
\newblock Mutan: Multimodal tucker fusion for visual question answering.
\newblock In \emph{Proceedings of the IEEE international conference on computer vision}, pages 2612--2620, 2017.

\bibitem[Bergmann et~al.(2021)Bergmann, Batzner, Fauser, Sattlegger, and Steger]{MVTec}
Paul Bergmann, Kilian Batzner, Michael Fauser, David Sattlegger, and Carsten Steger.
\newblock The mvtec anomaly detection dataset: a comprehensive real-world dataset for unsupervised anomaly detection.
\newblock \emph{International Journal of Computer Vision}, 129\penalty0 (4):\penalty0 1038--1059, 2021.

\bibitem[Brown et~al.(2020)Brown, Mann, Ryder, Subbiah, Kaplan, Dhariwal, Neelakantan, Shyam, Sastry, Askell, et~al.]{brown2020language}
Tom Brown, Benjamin Mann, Nick Ryder, Melanie Subbiah, Jared~D Kaplan, Prafulla Dhariwal, Arvind Neelakantan, Pranav Shyam, Girish Sastry, Amanda Askell, et~al.
\newblock Language models are few-shot learners.
\newblock \emph{Advances in neural information processing systems}, 33:\penalty0 1877--1901, 2020.

\bibitem[Cadene et~al.(2019)Cadene, Ben-Younes, Cord, and Thome]{cadene2019murel}
Remi Cadene, Hedi Ben-Younes, Matthieu Cord, and Nicolas Thome.
\newblock Murel: Multimodal relational reasoning for visual question answering.
\newblock In \emph{Proceedings of the IEEE/CVF conference on computer vision and pattern recognition}, pages 1989--1998, 2019.

\bibitem[Cao et~al.(2021)Cao, Wan, Wang, Liang, and Lin]{cao2021linguistically}
Qingxing Cao, Wentao Wan, Keze Wang, Xiaodan Liang, and Liang Lin.
\newblock Linguistically routing capsule network for out-of-distribution visual question answering.
\newblock In \emph{Proceedings of the IEEE/CVF International Conference on Computer Vision}, pages 1614--1623, 2021.

\bibitem[Chen et~al.(2023{\natexlab{a}})Chen, Zhu, Shen, Li, Liu, Zhang, Krishnamoorthi, Chandra, Xiong, and Elhoseiny]{chen2023minigptv2}
Jun Chen, Deyao Zhu, Xiaoqian Shen, Xiang Li, Zechun Liu, Pengchuan Zhang, Raghuraman Krishnamoorthi, Vikas Chandra, Yunyang Xiong, and Mohamed Elhoseiny.
\newblock Minigpt-v2: large language model as a unified interface for vision-language multi-task learning, 2023{\natexlab{a}}.

\bibitem[Chen et~al.(2023{\natexlab{b}})Chen, Zhang, Ren, Zhao, Cai, Wang, Wang, Liu, and Chang]{chen2023towards}
Liang Chen, Yichi Zhang, Shuhuai Ren, Haozhe Zhao, Zefan Cai, Yuchi Wang, Peiyi Wang, Tianyu Liu, and Baobao Chang.
\newblock Towards end-to-end embodied decision making via multi-modal large language model: Explorations with gpt4-vision and beyond.
\newblock \emph{arXiv preprint arXiv:2310.02071}, 2023{\natexlab{b}}.

\bibitem[Chen et~al.(2015)Chen, Fang, Lin, Vedantam, Gupta, Doll{\'a}r, and Zitnick]{chen2015microsoft}
Xinlei Chen, Hao Fang, Tsung-Yi Lin, Ramakrishna Vedantam, Saurabh Gupta, Piotr Doll{\'a}r, and C~Lawrence Zitnick.
\newblock Microsoft coco captions: Data collection and evaluation server.
\newblock \emph{arXiv preprint arXiv:1504.00325}, 2015.

\bibitem[Chowdhery et~al.(2022)Chowdhery, Narang, Devlin, Bosma, Mishra, Roberts, Barham, Chung, Sutton, Gehrmann, et~al.]{chowdhery2022palm}
Aakanksha Chowdhery, Sharan Narang, Jacob Devlin, Maarten Bosma, Gaurav Mishra, Adam Roberts, Paul Barham, Hyung~Won Chung, Charles Sutton, Sebastian Gehrmann, et~al.
\newblock Palm: Scaling language modeling with pathways.
\newblock \emph{arXiv preprint arXiv:2204.02311}, 2022.

\bibitem[Dai et~al.(2023)Dai, Li, Li, Tiong, Zhao, Wang, Li, Fung, and Hoi]{dai2023instructblip}
Wenliang Dai, Junnan Li, Dongxu Li, Anthony Meng~Huat Tiong, Junqi Zhao, Weisheng Wang, Boyang Li, Pascale Fung, and Steven Hoi.
\newblock Instructblip: Towards general-purpose vision-language models with instruction tuning, 2023.

\bibitem[Dancette et~al.(2021)Dancette, Cadene, Teney, and Cord]{dancette2021beyond}
Corentin Dancette, Remi Cadene, Damien Teney, and Matthieu Cord.
\newblock Beyond question-based biases: Assessing multimodal shortcut learning in visual question answering.
\newblock In \emph{Proceedings of the IEEE/CVF International Conference on Computer Vision}, pages 1574--1583, 2021.

\bibitem[Ding et~al.(2022)Ding, Yu, Liu, Hu, Cui, and Wu]{ding2022mukea}
Yang Ding, Jing Yu, Bang Liu, Yue Hu, Mingxin Cui, and Qi Wu.
\newblock Mukea: Multimodal knowledge extraction and accumulation for knowledge-based visual question answering.
\newblock In \emph{Proceedings of the IEEE/CVF Conference on Computer Vision and Pattern Recognition}, pages 5089--5098, 2022.

\bibitem[Fu et~al.(2023)Fu, Chen, Shen, Qin, Zhang, Lin, Qiu, Lin, Yang, Zheng, et~al.]{fu2023mme}
Chaoyou Fu, Peixian Chen, Yunhang Shen, Yulei Qin, Mengdan Zhang, Xu Lin, Zhenyu Qiu, Wei Lin, Jinrui Yang, Xiawu Zheng, et~al.
\newblock Mme: A comprehensive evaluation benchmark for multimodal large language models.
\newblock \emph{arXiv preprint arXiv:2306.13394}, 2023.

\bibitem[Fukui et~al.(2016)Fukui, Park, Yang, Rohrbach, Darrell, and Rohrbach]{fukui2016multimodal}
Akira Fukui, Dong~Huk Park, Daylen Yang, Anna Rohrbach, Trevor Darrell, and Marcus Rohrbach.
\newblock Multimodal compact bilinear pooling for visual question answering and visual grounding.
\newblock \emph{arXiv preprint arXiv:1606.01847}, 2016.

\bibitem[Gao et~al.(2022)Gao, Ping, Thattai, Reganti, Wu, and Natarajan]{gao2022transform}
Feng Gao, Qing Ping, Govind Thattai, Aishwarya Reganti, Ying~Nian Wu, and Prem Natarajan.
\newblock Transform-retrieve-generate: Natural language-centric outside-knowledge visual question answering.
\newblock In \emph{Proceedings of the IEEE/CVF Conference on Computer Vision and Pattern Recognition}, pages 5067--5077, 2022.

\bibitem[Ge et~al.(2023)Ge, Luo, Qian, Gan, Fu, and Zhan]{ge2023chain}
Jiaxin Ge, Hongyin Luo, Siyuan Qian, Yulu Gan, Jie Fu, and Shanghang Zhan.
\newblock Chain of thought prompt tuning in vision language models.
\newblock \emph{arXiv preprint arXiv:2304.07919}, 2023.

\bibitem[Goyal et~al.(2017)Goyal, Khot, Summers-Stay, Batra, and Parikh]{goyal2017making}
Yash Goyal, Tejas Khot, Douglas Summers-Stay, Dhruv Batra, and Devi Parikh.
\newblock Making the v in vqa matter: Elevating the role of image understanding in visual question answering.
\newblock In \emph{Proceedings of the IEEE conference on computer vision and pattern recognition}, pages 6904--6913, 2017.

\bibitem[Ha et~al.(2017)Ha, Watanabe, Karasawa, Ushiku, and Harada]{2017-infrared-vehicle}
Qishen Ha, Kohei Watanabe, Takumi Karasawa, Yoshitaka Ushiku, and Tatsuya Harada.
\newblock Mfnet: Towards real-time semantic segmentation for autonomous vehicles with multi-spectral scenes.
\newblock In \emph{2017 IEEE/RSJ International Conference on Intelligent Robots and Systems (IROS)}, pages 5108--5115. IEEE, 2017.

\bibitem[Han et~al.(2021)Han, Wang, Su, Huang, and Tian]{han2021greedy}
Xinzhe Han, Shuhui Wang, Chi Su, Qingming Huang, and Qi Tian.
\newblock Greedy gradient ensemble for robust visual question answering.
\newblock In \emph{Proceedings of the IEEE/CVF International Conference on Computer Vision}, pages 1584--1593, 2021.

\bibitem[Hudson and Manning(2019)]{hudson2019gqa}
Drew~A Hudson and Christopher~D Manning.
\newblock Gqa: A new dataset for real-world visual reasoning and compositional question answering.
\newblock In \emph{Proceedings of the IEEE/CVF conference on computer vision and pattern recognition}, pages 6700--6709, 2019.

\bibitem[Hwang et~al.(2015)Hwang, Park, Kim, Choi, and So~Kweon]{2015infra-pedestrian}
Soonmin Hwang, Jaesik Park, Namil Kim, Yukyung Choi, and In So~Kweon.
\newblock Multispectral pedestrian detection: Benchmark dataset and baseline.
\newblock In \emph{Proceedings of the IEEE conference on computer vision and pattern recognition}, pages 1037--1045, 2015.

\bibitem[Jia et~al.(2021)Jia, Zhu, Li, Tang, and Zhou]{2021infra-lowlight}
Xinyu Jia, Chuang Zhu, Minzhen Li, Wenqi Tang, and Wenli Zhou.
\newblock Llvip: A visible-infrared paired dataset for low-light vision.
\newblock In \emph{Proceedings of the IEEE/CVF international conference on computer vision}, pages 3496--3504, 2021.

\bibitem[Johnson et~al.(2017{\natexlab{a}})Johnson, Hariharan, Van Der~Maaten, Fei-Fei, Lawrence~Zitnick, and Girshick]{johnson2017clevr}
Justin Johnson, Bharath Hariharan, Laurens Van Der~Maaten, Li Fei-Fei, C Lawrence~Zitnick, and Ross Girshick.
\newblock Clevr: A diagnostic dataset for compositional language and elementary visual reasoning.
\newblock In \emph{Proceedings of the IEEE conference on computer vision and pattern recognition}, pages 2901--2910, 2017{\natexlab{a}}.

\bibitem[Johnson et~al.(2017{\natexlab{b}})Johnson, Hariharan, Van Der~Maaten, Hoffman, Fei-Fei, Lawrence~Zitnick, and Girshick]{johnson2017inferring}
Justin Johnson, Bharath Hariharan, Laurens Van Der~Maaten, Judy Hoffman, Li Fei-Fei, C Lawrence~Zitnick, and Ross Girshick.
\newblock Inferring and executing programs for visual reasoning.
\newblock In \emph{Proceedings of the IEEE international conference on computer vision}, pages 2989--2998, 2017{\natexlab{b}}.

\bibitem[Kim et~al.(2016)Kim, On, Lim, Kim, Ha, and Zhang]{kim2016hadamard}
Jin-Hwa Kim, Kyoung-Woon On, Woosang Lim, Jeonghee Kim, Jung-Woo Ha, and Byoung-Tak Zhang.
\newblock Hadamard product for low-rank bilinear pooling.
\newblock \emph{arXiv preprint arXiv:1610.04325}, 2016.

\bibitem[Kim et~al.(2018)Kim, Jun, and Zhang]{kim2018bilinear}
Jin-Hwa Kim, Jaehyun Jun, and Byoung-Tak Zhang.
\newblock Bilinear attention networks.
\newblock \emph{Advances in neural information processing systems}, 31, 2018.

\bibitem[Li et~al.(2023{\natexlab{a}})Li, Wang, Wang, Ge, Ge, and Shan]{li2023seed}
Bohao Li, Rui Wang, Guangzhi Wang, Yuying Ge, Yixiao Ge, and Ying Shan.
\newblock Seed-bench: Benchmarking multimodal llms with generative comprehension.
\newblock \emph{arXiv preprint arXiv:2307.16125}, 2023{\natexlab{a}}.

\bibitem[Li et~al.(2023{\natexlab{b}})Li, Zhang, Chen, Wang, Yang, and Liu]{li2023otter}
Bo Li, Yuanhan Zhang, Liangyu Chen, Jinghao Wang, Jingkang Yang, and Ziwei Liu.
\newblock Otter: A multi-modal model with in-context instruction tuning.
\newblock \emph{arXiv preprint arXiv:2305.03726}, 2023{\natexlab{b}}.

\bibitem[Li et~al.(2023{\natexlab{c}})Li, Du, Zhou, Wang, Zhao, and Wen]{li2023evaluating}
Yifan Li, Yifan Du, Kun Zhou, Jinpeng Wang, Wayne~Xin Zhao, and Ji-Rong Wen.
\newblock Evaluating object hallucination in large vision-language models.
\newblock \emph{arXiv preprint arXiv:2305.10355}, 2023{\natexlab{c}}.

\bibitem[Li et~al.(2023{\natexlab{d}})Li, Du, Zhou, Wang, Zhao, and Wen]{li2023pope}
Yifan Li, Yifan Du, Kun Zhou, Jinpeng Wang, Wayne~Xin Zhao, and Ji-Rong Wen.
\newblock Evaluating object hallucination in large vision-language models.
\newblock \emph{arXiv preprint arXiv:2305.10355}, 2023{\natexlab{d}}.

\bibitem[Liang et~al.(2021)Liang, Lyu, Fan, Wu, Cheng, Wu, Chen, Wu, Lee, Zhu, et~al.]{liang2021multibench}
Paul~Pu Liang, Yiwei Lyu, Xiang Fan, Zetian Wu, Yun Cheng, Jason Wu, Leslie Chen, Peter Wu, Michelle~A Lee, Yuke Zhu, et~al.
\newblock Multibench: Multiscale benchmarks for multimodal representation learning.
\newblock \emph{arXiv preprint arXiv:2107.07502}, 2021.

\bibitem[Liu et~al.(2021)Liu, Zhan, Xu, Ma, Yang, and Wu]{isbi}
Bo Liu, Li-Ming Zhan, Li Xu, Lin Ma, Yan Yang, and Xiao-Ming Wu.
\newblock Slake: A semantically-labeled knowledge-enhanced dataset for medical visual question answering.
\newblock In \emph{2021 IEEE 18th International Symposium on Biomedical Imaging (ISBI)}, pages 1650--1654. IEEE, 2021.

\bibitem[Liu et~al.(2023{\natexlab{a}})Liu, Li, Li, and Lee]{liu2023improved}
Haotian Liu, Chunyuan Li, Yuheng Li, and Yong~Jae Lee.
\newblock Improved baselines with visual instruction tuning.
\newblock \emph{arXiv preprint arXiv:2310.03744}, 2023{\natexlab{a}}.

\bibitem[Liu et~al.(2023{\natexlab{b}})Liu, Li, Wu, and Lee]{liu2023visual}
Haotian Liu, Chunyuan Li, Qingyang Wu, and Yong~Jae Lee.
\newblock Visual instruction tuning.
\newblock \emph{arXiv preprint arXiv:2304.08485}, 2023{\natexlab{b}}.

\bibitem[Liu et~al.(2022)Liu, Fan, Huang, Wu, Liu, Zhong, and Luo]{2022-infrared-detect}
Jinyuan Liu, Xin Fan, Zhanbo Huang, Guanyao Wu, Risheng Liu, Wei Zhong, and Zhongxuan Luo.
\newblock Target-aware dual adversarial learning and a multi-scenario multi-modality benchmark to fuse infrared and visible for object detection.
\newblock In \emph{Proceedings of the IEEE/CVF Conference on Computer Vision and Pattern Recognition (CVPR)}, pages 5802--5811, 2022.

\bibitem[Liu et~al.(2023{\natexlab{c}})Liu, Duan, Zhang, Li, Zhang, Zhao, Yuan, Wang, He, Liu, et~al.]{liu2023mmbench}
Yuan Liu, Haodong Duan, Yuanhan Zhang, Bo Li, Songyang Zhang, Wangbo Zhao, Yike Yuan, Jiaqi Wang, Conghui He, Ziwei Liu, et~al.
\newblock Mmbench: Is your multi-modal model an all-around player?
\newblock \emph{arXiv preprint arXiv:2307.06281}, 2023{\natexlab{c}}.

\bibitem[Manyika(2023)]{manyika2023overview}
James Manyika.
\newblock An overview of bard: an early experiment with generative ai.
\newblock \emph{AI. Google Static Documents}, 2023.

\bibitem[Mao et~al.(2023)Mao, Chen, Zhu, Chen, Su, Zhang, and Xue]{mao2023coco}
Xiaofeng Mao, Yuefeng Chen, Yao Zhu, Da Chen, Hang Su, Rong Zhang, and Hui Xue.
\newblock Coco-o: A benchmark for object detectors under natural distribution shifts.
\newblock In \emph{Proceedings of the IEEE/CVF International Conference on Computer Vision}, pages 6339--6350, 2023.

\bibitem[Marino et~al.(2019)Marino, Rastegari, Farhadi, and Mottaghi]{marino2019ok}
Kenneth Marino, Mohammad Rastegari, Ali Farhadi, and Roozbeh Mottaghi.
\newblock Ok-vqa: A visual question answering benchmark requiring external knowledge.
\newblock In \emph{Proceedings of the IEEE/cvf conference on computer vision and pattern recognition}, pages 3195--3204, 2019.

\bibitem[Nam et~al.(2017)Nam, Ha, and Kim]{nam2017dual}
Hyeonseob Nam, Jung-Woo Ha, and Jeonghee Kim.
\newblock Dual attention networks for multimodal reasoning and matching.
\newblock In \emph{Proceedings of the IEEE conference on computer vision and pattern recognition}, pages 299--307, 2017.

\bibitem[Nguyen et~al.(2022)Nguyen, Cai, Zhao, Kot, and Wen]{nguyen2022towards}
Hong~Duc Nguyen, Rizhao Cai, Heng Zhao, Alex~C Kot, and Bihan Wen.
\newblock Towards more efficient security inspection via deep learning: a task-driven x-ray image cropping scheme.
\newblock \emph{Micromachines}, 13\penalty0 (4):\penalty0 565, 2022.

\bibitem[OpenAI(2023)]{openai2023gpt}
R OpenAI.
\newblock Gpt-4 technical report. arxiv 2303.08774.
\newblock \emph{View in Article}, 2023.

\bibitem[Shah et~al.(2019)Shah, Chen, Rohrbach, and Parikh]{shah2019cycle}
Meet Shah, Xinlei Chen, Marcus Rohrbach, and Devi Parikh.
\newblock Cycle-consistency for robust visual question answering.
\newblock In \emph{Proceedings of the IEEE/CVF Conference on Computer Vision and Pattern Recognition}, pages 6649--6658, 2019.

\bibitem[Shi et~al.(2020)Shi, Zhou, Qiu, and Zhu]{shi2020improving}
Zhan Shi, Xu Zhou, Xipeng Qiu, and Xiaodan Zhu.
\newblock Improving image captioning with better use of captions.
\newblock \emph{arXiv preprint arXiv:2006.11807}, 2020.

\bibitem[Sidorov et~al.(2020)Sidorov, Hu, Rohrbach, and Singh]{sidorov2020textcaps}
Oleksii Sidorov, Ronghang Hu, Marcus Rohrbach, and Amanpreet Singh.
\newblock Textcaps: a dataset for image captioning with reading comprehension.
\newblock In \emph{Computer Vision--ECCV 2020: 16th European Conference, Glasgow, UK, August 23--28, 2020, Proceedings, Part II 16}, pages 742--758. Springer, 2020.

\bibitem[Singh et~al.(2019)Singh, Natarajan, Shah, Jiang, Chen, Batra, Parikh, and Rohrbach]{singh2019towards}
Amanpreet Singh, Vivek Natarajan, Meet Shah, Yu Jiang, Xinlei Chen, Dhruv Batra, Devi Parikh, and Marcus Rohrbach.
\newblock Towards vqa models that can read.
\newblock In \emph{Proceedings of the IEEE/CVF conference on computer vision and pattern recognition}, pages 8317--8326, 2019.

\bibitem[Tan and Bansal(2019)]{tan2019lxmert}
Hao Tan and Mohit Bansal.
\newblock Lxmert: Learning cross-modality encoder representations from transformers.
\newblock \emph{arXiv preprint arXiv:1908.07490}, 2019.

\bibitem[Tascon-Morales et~al.(2023)Tascon-Morales, M{\'a}rquez-Neila, and Sznitman]{tascon2023logical}
Sergio Tascon-Morales, Pablo M{\'a}rquez-Neila, and Raphael Sznitman.
\newblock Logical implications for visual question answering consistency.
\newblock In \emph{Proceedings of the IEEE/CVF Conference on Computer Vision and Pattern Recognition}, pages 6725--6735, 2023.

\bibitem[Teney et~al.(2021)Teney, Abbasnejad, and van~den Hengel]{teney2021unshuffling}
Damien Teney, Ehsan Abbasnejad, and Anton van~den Hengel.
\newblock Unshuffling data for improved generalization in visual question answering.
\newblock In \emph{Proceedings of the IEEE/CVF international conference on computer vision}, pages 1417--1427, 2021.

\bibitem[Touvron et~al.(2023)Touvron, Lavril, Izacard, Martinet, Lachaux, Lacroix, Rozi{\`e}re, Goyal, Hambro, Azhar, et~al.]{touvron2023llama}
Hugo Touvron, Thibaut Lavril, Gautier Izacard, Xavier Martinet, Marie-Anne Lachaux, Timoth{\'e}e Lacroix, Baptiste Rozi{\`e}re, Naman Goyal, Eric Hambro, Faisal Azhar, et~al.
\newblock Llama: Open and efficient foundation language models.
\newblock \emph{arXiv preprint arXiv:2302.13971}, 2023.

\bibitem[Tsimpoukelli et~al.(2021)Tsimpoukelli, Menick, Cabi, Eslami, Vinyals, and Hill]{tsimpoukelli2021multimodal}
Maria Tsimpoukelli, Jacob~L Menick, Serkan Cabi, SM Eslami, Oriol Vinyals, and Felix Hill.
\newblock Multimodal few-shot learning with frozen language models.
\newblock \emph{Advances in Neural Information Processing Systems}, 34:\penalty0 200--212, 2021.

\bibitem[Vaswani et~al.(2017)Vaswani, Shazeer, Parmar, Uszkoreit, Jones, Gomez, Kaiser, and Polosukhin]{vaswani2017attention}
Ashish Vaswani, Noam Shazeer, Niki Parmar, Jakob Uszkoreit, Llion Jones, Aidan~N Gomez, {\L}ukasz Kaiser, and Illia Polosukhin.
\newblock Attention is all you need.
\newblock \emph{Advances in neural information processing systems}, 30, 2017.

\bibitem[Xia et~al.(2018)Xia, Bai, Ding, Zhu, Belongie, Luo, Datcu, Pelillo, and Zhang]{xia2018dota}
Gui-Song Xia, Xiang Bai, Jian Ding, Zhen Zhu, Serge Belongie, Jiebo Luo, Mihai Datcu, Marcello Pelillo, and Liangpei Zhang.
\newblock Dota: A large-scale dataset for object detection in aerial images.
\newblock In \emph{Proceedings of the IEEE conference on computer vision and pattern recognition}, pages 3974--3983, 2018.

\bibitem[Xu et~al.(2023)Xu, Shao, Zhang, Gao, Liu, Lei, Meng, Huang, Qiao, and Luo]{xu2023lvlm}
Peng Xu, Wenqi Shao, Kaipeng Zhang, Peng Gao, Shuo Liu, Meng Lei, Fanqing Meng, Siyuan Huang, Yu Qiao, and Ping Luo.
\newblock Lvlm-ehub: A comprehensive evaluation benchmark for large vision-language models.
\newblock \emph{arXiv preprint arXiv:2306.09265}, 2023.

\bibitem[Yang et~al.(2021)Yang, Gao, Zhang, and Cai]{yang2021auto}
Xu Yang, Chongyang Gao, Hanwang Zhang, and Jianfei Cai.
\newblock Auto-parsing network for image captioning and visual question answering.
\newblock In \emph{Proceedings of the IEEE/CVF International Conference on Computer Vision}, pages 2197--2207, 2021.

\bibitem[Yang et~al.(2023{\natexlab{a}})Yang, Li, Lin, Wang, Lin, Liu, and Wang]{yang2023dawn}
Zhengyuan Yang, Linjie Li, Kevin Lin, Jianfeng Wang, Chung-Ching Lin, Zicheng Liu, and Lijuan Wang.
\newblock The dawn of lmms: Preliminary explorations with gpt-4v (ision).
\newblock \emph{arXiv preprint arXiv:2309.17421}, 2023{\natexlab{a}}.

\bibitem[Yang et~al.(2023{\natexlab{b}})Yang, Li, Wang, Lin, Azarnasab, Ahmed, Liu, Liu, Zeng, and Wang]{yang2023mm}
Zhengyuan Yang, Linjie Li, Jianfeng Wang, Kevin Lin, Ehsan Azarnasab, Faisal Ahmed, Zicheng Liu, Ce Liu, Michael Zeng, and Lijuan Wang.
\newblock Mm-react: Prompting chatgpt for multimodal reasoning and action.
\newblock \emph{arXiv preprint arXiv:2303.11381}, 2023{\natexlab{b}}.

\bibitem[Ye et~al.(2023)Ye, Xu, Xu, Ye, Yan, Zhou, Wang, Hu, Shi, Shi, et~al.]{ye2023mplug}
Qinghao Ye, Haiyang Xu, Guohai Xu, Jiabo Ye, Ming Yan, Yiyang Zhou, Junyang Wang, Anwen Hu, Pengcheng Shi, Yaya Shi, et~al.
\newblock mplug-owl: Modularization empowers large language models with multimodality.
\newblock \emph{arXiv preprint arXiv:2304.14178}, 2023.

\bibitem[Yu et~al.(2020)Yu, Chen, Wang, Xian, Chen, Liu, Madhavan, and Darrell]{BDD100K}
Fisher Yu, Haofeng Chen, Xin Wang, Wenqi Xian, Yingying Chen, Fangchen Liu, Vashisht Madhavan, and Trevor Darrell.
\newblock Bdd100k: A diverse driving dataset for heterogeneous multitask learning.
\newblock In \emph{2020 IEEE/CVF Conference on Computer Vision and Pattern Recognition (CVPR)}, pages 2633--2642, 2020.

\bibitem[Yu et~al.(2022)Yu, Wu, Liang, Salakhutdinov, and Morency]{yu2022pacs}
Samuel Yu, Peter Wu, Paul~Pu Liang, Ruslan Salakhutdinov, and Louis-Philippe Morency.
\newblock Pacs: A dataset for physical audiovisual commonsense reasoning.
\newblock In \emph{European Conference on Computer Vision}, pages 292--309. Springer, 2022.

\bibitem[Yu et~al.(2023)Yu, Yang, Li, Wang, Lin, Liu, Wang, and Wang]{yu2023mmvet}
Weihao Yu, Zhengyuan Yang, Linjie Li, Jianfeng Wang, Kevin Lin, Zicheng Liu, Xinchao Wang, and Lijuan Wang.
\newblock Mm-vet: Evaluating large multimodal models for integrated capabilities.
\newblock \emph{arXiv preprint arXiv:2308.02490}, 2023.

\bibitem[Zellers et~al.(2019)Zellers, Bisk, Farhadi, and Choi]{zellers2019recognition}
Rowan Zellers, Yonatan Bisk, Ali Farhadi, and Yejin Choi.
\newblock From recognition to cognition: Visual commonsense reasoning.
\newblock In \emph{Proceedings of the IEEE/CVF conference on computer vision and pattern recognition}, pages 6720--6731, 2019.

\bibitem[Zhang et~al.(2023)Zhang, Zhang, Li, Zhao, Karypis, and Smola]{zhang2023multimodal}
Zhuosheng Zhang, Aston Zhang, Mu Li, Hai Zhao, George Karypis, and Alex Smola.
\newblock Multimodal chain-of-thought reasoning in language models.
\newblock \emph{arXiv preprint arXiv:2302.00923}, 2023.

\bibitem[Zhou et~al.(2021)Zhou, Ren, Zhu, Sun, Liu, Ding, Xu, and Ji]{zhou2021trar}
Yiyi Zhou, Tianhe Ren, Chaoyang Zhu, Xiaoshuai Sun, Jianzhuang Liu, Xinghao Ding, Mingliang Xu, and Rongrong Ji.
\newblock Trar: Routing the attention spans in transformer for visual question answering.
\newblock In \emph{Proceedings of the IEEE/CVF International Conference on Computer Vision}, pages 2074--2084, 2021.

\bibitem[Zhu et~al.(2023)Zhu, Chen, Shen, Li, and Elhoseiny]{zhu2023minigpt4_v1}
Deyao Zhu, Jun Chen, Xiaoqian Shen, Xiang Li, and Mohamed Elhoseiny.
\newblock Minigpt-4: Enhancing vision-language understanding with advanced large language models, 2023.

\end{thebibliography}
}

\end{document}